\documentclass[10pt,twocolumn,letterpaper]{article}

\usepackage{cvpr}

\makeatletter
\@namedef{ver@everyshi.sty}{}
\makeatother
\usepackage{tikz}

\usepackage{times}
\usepackage{epsfig}
\usepackage{graphicx}
\usepackage{amsmath}
\usepackage{amssymb}
\usepackage{xspace}
\usepackage{balance}

\usepackage[pagebackref=true,breaklinks=true,colorlinks,bookmarks=false]{hyperref}

\newcounter{Lcount}
\newcommand{\numsquishlist}{
   \begin{list}{\arabic{Lcount}. }
    { \usecounter{Lcount}
 \setlength{\itemsep}{-.1ex}      \setlength{\parsep}{0ex}
      \setlength{\topsep}{0ex}       \setlength{\partopsep}{0ex}
      \setlength{\leftmargin}{1em} \setlength{\labelwidth}{1em}
      \setlength{\labelsep}{0.1em} } }
\newcommand{\numsquishend}{\end{list}}

\newcommand{\squishlist}{
   \begin{list}{$\bullet$}
    { \setlength{\itemsep}{-.1ex}      \setlength{\parsep}{0ex}
      \setlength{\topsep}{0ex}       \setlength{\partopsep}{0ex}
      \setlength{\leftmargin}{.8em} \setlength{\labelwidth}{1em}
      \setlength{\labelsep}{0.5em} } }
\newcommand{\squishend}{\end{list}}

\cvprfinalcopy 

\def\cvprPaperID{31} 

\newcommand{\cc}{{\sc Continual Curation}\xspace}

\ifcvprfinal\fi

\begin{document}

\title{The Animal ID Problem: Continual Curation}

\author{C.\ V.\ Stewart\\
Rensselaer Polytechnic Institute \\
{\tt\small stewart@rpi.edu}
\and
J.\ R.\ Parham, J.\ Holmberg \\
Wild Me \\
{\tt\small \{parham,jason\}@wildme.org}
\and
T.\ Y.\ Berger-Wolf \\
Ohio State University \\
{\tt\small berger-wolf.1@osu.edu}
}

\maketitle

\begin{abstract}
Hoping to stimulate new research in individual animal identification from images, we propose to formulate the problem as the human-machine {\sc Continual Curation} of images and animal identities.  
This is an open world recognition problem 
\cite{bendale2015towards}, where most new animals enter the system after its algorithms are initially trained and deployed.
\cc, as defined here, requires (1) an improvement in the effectiveness of current recognition methods, (2) a pairwise verification algorithm that allows the possibility of no decision, and (3) an algorithmic decision mechanism that seeks human input to guide the curation process.  Error metrics must evaluate the ability of recognition algorithms to identify not only animals that have been seen just once or twice but also recognize new animals not in the database.  An important measure of overall system performance is accuracy as a function of the amount of human input required.
\end{abstract}

\section{Introduction}

The challenging problem of automatically identifying unique animals from images (animal ID) is of growing importance in biology, ecology and conservation.  See \cite{vidal2021perspectives} for a recent review.
Applications in animal ID range from population surveys to studying the behavior and movements of single animals.
This problem has received more attention by computer vision community as well, as techniques have been developed for a wide variety of species, including small birds \cite{ferreira-bird-id-2020}, penguins, primates \cite{deb2018face, freytagChimpanzeeFacesWild2016, schofield2019chimpanzee}, ungulates \cite{crall2013hotspotter}, large predators \cite{clapham-bear-face-2020, liAmurTigerReidentification2019}, elephants \cite{Weideman_2020_WACV}, manta rays \cite{moskvyak2019robust} sharks \cite{arzoumanian2005astronomical, hughes2017automated} and whales \cite{Weideman_2020_WACV}.

Most work in the computer vision literature on animal ID treats the problem as an extension of the object and face recognition problem.
In particular, the database of individuals is assumed to largely be closed, the problem is posed as one of retrieval, the primary goal is rank-ordering of potential IDs for each query, and few experiments are presented on handling novel identities.
While this problem setting is important for some applications -- and improvements in retrieval algorithm performance are still needed -- we argue that a broader formulation, to which we refer as \cc, is needed for the animal ID problem. This is a problem where (1) most animals are added to the system after it is initially trained and deployed, (2) when these new animals are added it is not known that they are truly novel, and (3) individuals often must be re-identified very soon (if not immediately) after being added to the system. 

Our formulation has a number of important implications for the development of animal ID algorithms that we discuss in this paper.  We start by defining the problem in detail (Section~\ref{sec:continual}).  We then outline the algorithm pipeline we are developing for the 
REDACTED\footnote{\label{redact-wildbook}Name of author's public animal ID product redacted for peer review.}
system, and consider its relationship to associated problems and techniques in machine learning and computer vision (Section~\ref{sec:approaches}).  Finally, we introduce several performance metrics (Section~\ref{sec:performance}).  While our system remains a work in progress, our goal is to stimulate new work in the field, leading to the development and sharing of datasets, algorithms, performance metrics and software throughout the community.

Before beginning, we must stress that our focus is on algorithmic issues. There remain equally important application issues surrounding the collection, protection and sharing of animal ID data and its derived metadata, especially for endangered or poached species.  For brevity and focus, we leave those considerations for future discussions.

\section{Animal ID and Continual Curation} \label{sec:continual}

The following observations about animal ID are based on real-world experience in using prototype systems for animal population monitoring and management. 

\begin{figure*}[t]
\begin{center}
   \includegraphics[width=1.0\linewidth]{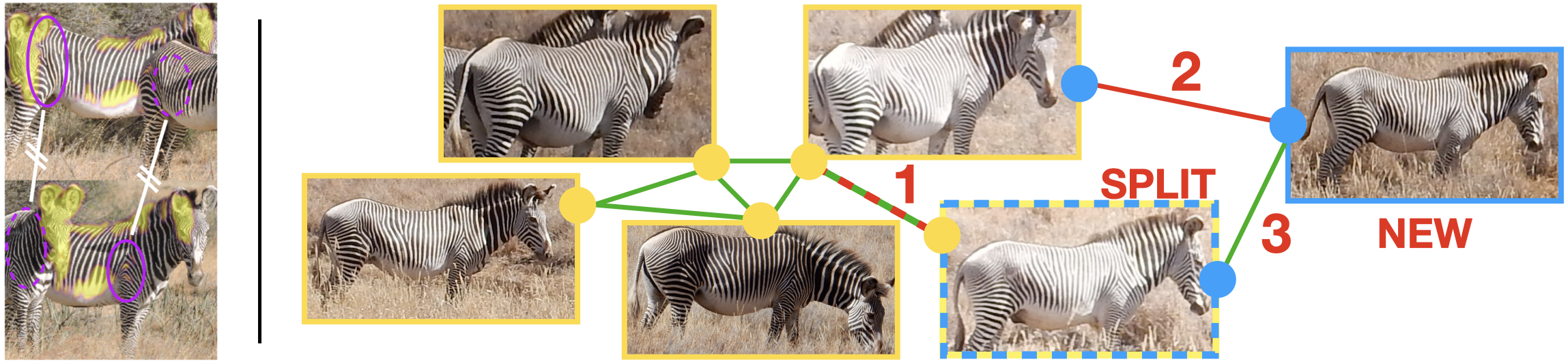}
\end{center}
   \caption{[LEFT] An example of two incomparable annotations of the same animal. [RIGHT] An example of a split decision.  A database consists of 5 annotations for 1 ID (yellow) with positive matches.  new image (blue) is added and gets both negative (2) and positive (3) decisions.  To fix the problem, a split is done by making (1) negative and adding the blue/yellow annotation to the blue ID.}
   \label{combined}
\end{figure*}

\numsquishlist
\item New individuals are continuously added to the system, making animal ID an open world recognition problem \cite{bendale2015towards}).  Moreover, when new images are added it is not known which of them may show new individuals. Thus, any solution must recognize when an individual is new and must \textit{quickly} learn to recognize it in subsequent images. These new individuals arrive in an unpredictable pattern.
Thus, recognition of the new individual may be based on only one or a few examples. This is a fundamental challenge in recognition (\cite{scheirer2012toward}).
\item The distribution of animal sightings is typically heavily skewed, with many individuals appearing once in the database. For example, in a large whale shark dataset about 45\% of over 10,000 individuals were seen once, while in a large Grevy's zebra census, 31\% of over 2,000 individuals were only seen once.
\item Humans are an inherent part of the curation process,  as training data annotators as well as the ultimate reviewers. Moreover, it is expected and often required that the curation decisions will include human judgement.
\item Humans typically find it challenging, given an image, to identify the animal in the image from a large animal population. (This is not necessarily true of small populations like the few rhinos on a conservancy.)  Instead, humans are good at looking at two images and determining whether or not they show the same animal. The implications of this are quite important. Fundamentally, working without intelligent tools, a human would need to compare each query image to a representative image from each individual animal to decide whether it is new. The manual effort therefore grows quadratically with population size, leading to limits on the size of training sets or accuracy. 
\item Metadata about IDs is rarely available, especially compared to face recognition datasets where identifying information can be gleaned from government records, surrounding text, and social media content \cite{kemelmacher-shlizermanMegaFaceBenchmarkMillion2016}.
\item Mistakes will enter the curation results. These are naturally caused by both imperfect algorithms and fallible human reviewers, but this does not tell the whole story. The vagaries of the data are often an important part of the issue. For example, minor differences in viewpoint and lighting may make two pictures that show the same individual look different. A third image may eventually be introduced that matches reliably against each. Animal ID algorithms must discover this and adjust, leading to the merge of previously distinct animals. Conversely, adding a new image may uncover the need to ``split'' of an animal in the database.  See Figure \ref{combined} (right).
\item Not all animals depicted in an image are identifiable.  This can be caused by the pose of the animal, occlusions from vegetation or other animals, poor image quality, or in sufficient resolution.
See Figure \ref{combined} (left).  \textit{Identifiability} is not an absolute distinction, however, meaning that different humans and algorithms may produce different ID decisions for the same image.
\item There is an inherent trade-off across the effort of manual curation, the completeness of data, and the accuracy of curation \cite{beard_ms_2020}. If achieving near-perfect curation at all times is a central goal then the required human effort must increase to compensate for any potential failures of the underlying algorithms. Placing stricter controls — through preliminary detection algorithms or manual filtering — on which images become candidates for identification, could allow algorithms to perform better statistically and reduce the human effort. This is appropriate if rapid, low-cost population censusing is the goal. On the other hand, these controls may be loosened if extracting social networks or tracking few individuals is the goal.
\item
In part due to the limitations of human annotators, when dealing with moderate sized populations, most individuals will enter the system as ``new'' rather than as part of an initial training set. Thus, there is no real sense of a training phase based on curated data followed by a testing phase where the system is used.
\numsquishend
These factors lead us to an expanded view of the animal ID problem as \cc, where, starting from an often-small initial training set, each new set of images added to the database can show a new animal, can lead to the discovery and correction of mistakes, may require a series of  human decisions, and will require the system to adjust itself for the next round of additions.  More formally, we define the problem of \cc as, given a stream of images from an open set of animal identities, the goal is to continuously maintain the identification of animals in the images, allowing for human decisions.

\section{Ideas and Approaches} \label{sec:approaches}

We are designing the next generation of REDACTED to address individual animal identification as a \cc  problem. This section describes the major components of the design and considers related work in computer vision and machine learning.  This latter discussion is likely incomplete and we look forward to suggestions and discussion from the community.

\subsection{Components of a Solution}

Processing starts with a set of one or more query images and a database of images, annotations, and relationships. An \emph{annotation} is the region surrounding a detected individual in an image, together with a species label and other attributes. \textit{Relationships} between annotations indicate whether or not they show the same individual.
The database, including the relationships between annotations, is considered dynamic and may change at any time. 

The following are the processing steps for sets of query images:
\numsquishlist
    \item \textbf{Detection}: Each image is processed to find annotations. Instance segmentation may be applied as well where sufficient training data are available. Details of this step, while important, are not considered here, but see \cite{parhamAnimalDetectionPipeline2018}.
    \item \textbf{Filtering}: Annotations are filtered according to species, quality, and viewpoint. The goal is to ensure that distinguishing information appears in each annotation.  They may be filtered further to ensure that they all show roughly the same coverage of the distinguishing features -- e.g. both shoulder and hip on a zebra.
    \item \textbf{Ranking}: each query annotation is matched against other query annotations and against the database to produce potential matching animals. (Each query annotation starts with its own unique, temporary name.)  These are then rank-ordered, giving output typical of an object recognition or human face ID algorithm.
    

    \item \textbf{Verification}: Two potentially matching annotations are  evaluated to determine whether they show the same animal, show different animals, or there is insufficient information to tell -- for example if they show non-overlapping views of distinguishing features. This ``incomparable'' label is especially important when there are few annotations per individual, avoiding premature mistakes, allowing tradeoff between accuracy and uncertainty. Adding the incomparable label means that verification decisions must be based on more than just examining distances in a latent embedding space since distances for incomparable pairs of annotations are unlikely to be meaningful, especially when there are few annotations for an individual.
    \item \textbf{Decision}: the information provided by the detection, filtering, ranking, and verification algorithms must be combined into identity decisions. Given the imperfect nature of the algorithms, human input is needed as a guide. The human could be given complete control starting from the ranking, which is appropriate for smaller, closed populations.  For larger populations and more frequent queries, some level of automation is necessary to coordinate information and discover inconsistencies.  We treat the problem as one of dynamic, interactive clustering \cite{bae2020interactive}, where each cluster should contain all annotations from a single animal, but these clustering decisions can change dynamically as new input is provided. An important consideration is to determine what new information to seek.  Based on the discussion above, for humans this must be in the form of decisions about whether or not two annotations (or a small set of annotations) show the same animal. In essence this functions as inserting a human as an alternative to algorithmic verification algorithms, particularly under uncertainty.
\numsquishend

\subsection{Relation to Common Problems in Machine Learning and Computer Vision}

The following is a brief discussion of the overlap between the \cc problem and the relevant computer vision and machine learning literature.
The detection problem dovetails nicely with work on the broader detection problem, with an emphasis on overlap, viewpoint, quality, and occlusion \cite{parhamAnimalDetectionPipeline2018}. For filtering, characterizing identifiability is the most significant open question.

In ranking and verification, 
the most promising direction -- and the one we are exploring -- is semi-supervised and self-supervised learning.
Recent work has shown marked improvements in standard classification problems through the use of contrastive loss together with heavily augmented positive samples and simultaneous optimization over large numbers of negative samples \cite{chen2020simple, chen2020big, chen2020improved, He_2020_CVPR}.
This has been extended to a combination of supervised and unsupervised learning, which has shown experimental performance improvements over either alone \cite{khosla2020supervised}.
On the other hand, continued work on backbone architectures and training loss functions does not seem poised for breakthroughs of significance to animal ID.
For example, experimental evidence in \cite{musgrave2020metric}
show minimal performance differences between different loss functions once the backbone architecture is fixed.

Other directions of investigation are important for the ranking step. On a surprising number of species, traditional methods based on keypoint matching have proven effective \cite{crall2013hotspotter}.
In many cases they may be used to bootstrap training data for deep learning recognition algorithms.
Alternatively, keypoint-based methods may be applied in combination with deep learning methods where such methods are least-likely to succeed: novel individuals and few sightings \cite{DBLP:journals/corr/abs-1902-09324}.
Also interestingly is the idea of learning features from annotations without ID labels.
This has shown success in learning to detect dorsal fin edges on dolphins and other cetaceans, humpback flukes and elephant ears \cite{Weideman_2020_WACV}. Descriptors for matching are then computed from outlines.  Extracting training data requires significantly more work per image than bounding box labeling, but avoids the combinatorial problem of manual matching and, in principle, can even be applied where no IDs are available.

The verification problem posed here is a twist on traditional verification problems -- most notably for face ID \cite{papesh2018photo} -- because of our introduction of the ``incomparable'' option as a three-way classification problem. Interestingly, high-confidence negative decisions may provide additional data input for contrastive loss functions, or help reduce the number of verification pairs a human must examine.

Finally, the decision problem is perhaps most closely related to interactive clustering \cite{bae2020interactive}.
The key addition to the problem is that algorithmic guidance is needed to make decisions about which annotations to show to humans, focusing the human effort.  The alternative, typical of interactive clustering, is manual inspection of all results.  This is labor intensive and does not scale with the size of the datasets.

\section{Developing Performance Metrics} \label{sec:performance}

In suggesting ideas for experimental protocols for \cc, we focus on three areas: the performance of the ranking and verification algorithms, the overall system performance, and the amount of human effort.  We must work from datasets that have already been curated, and re-create or simulate the conditions under which they were added to the system.  (As our tools develop we will be able to curate increasingly larger sets.)  Given  $M$  annotations, we assume $m$ are already labeled and the remaining $M-m$ are added.

Measuring the performance of a verification algorithm is the most straightforward, requiring only that we have sufficient examples to handle the expected imbalance between positive, negative, and incomparable pairs.  Judgment of incomparable is a human decision.  Evaluating performance for ranking and for latent space embedding requires splitting the data so that among the $M-m$ test annotations some individuals are modeled as new, some individuals have been added since training completed, and the rest were used in training.  Important for the latter two is the number of times an animal has previously been seen.


Turning to the judgment of overall performance for identifying $M$ annotations, we work with clustering measures, where a cluster is a group of annotations determined to be the same animal. We assume a relatively large number of clusters, most or which are relatively small ($O(1)$ in size). There is wide variety of clustering performance measures, including precision and recall.
These measures are dominated by large clusters.  A simpler measure, well-suited to small clusters, is the number of extracted clusters that are also a ground-truth clusters, and the number of ground-truth clusters that are also found in the extracted clusters.  These are measures of whether the clusters correctly represent the true animal ID.  We propose the geometric mean of these two as a summary statistic.  A final simple measure of accuracy is the actual number of clusters extracted.  This is appropriate for population surveys where the primary goal is counting.

Human decisions are necessary to bridge the gap between the results of imperfect identification algorithms and accurate curation, while engendering trust in the engineered systems.  Therefore, one important measure is the level of accuracy as a function of human effort.  The primary human effort we propose to measure is the number of verification decisions made manually. Since it is impractical to involve human decision-making in repeated experiments involving thousands of images, we suggest a simulation: algorithms request human verification decisions from an oracle, which consults the ground truth data and returns the correct answer with high probability and an intentionally incorrect answer otherwise.
By varying this probability we can analyze both the impact and the tradeoff of human effort and accuracy.

These measures are straightforward to implement, but an important challenge in using them is having sufficient ID data.
There are some species such as zebras, humpback whales, whale sharks and others\footnote{LILA BC - \url{http://lila.science/datasets/}} for which this may be possible and we hope to release datasets in the future.
We also encourage the community to consider ways to contribute datasets for the study of \cc.

\section{Discussion}

We have presented an expanded view of the animal ID problem that we refer to as \cc.  It is an open world problem featuring frequent addition of unknown new individuals through query images. It also features skewed distributions of animal sightings with many animals seen only once or just a few times, yet must be recognized in future images. These present significant challenges to current recognition technologies.  They also necessitate  an algorithmic decision-making mechanism based on interactive clustering. Given the challenges of the problem, and the need for accurate results, human decisions must be an integral part of the solution. In summary, we propose to measure overall system effectiveness by the accuracy of curation as a function of human effort.

Important next steps include implementation of the proposed experimental protocols and performance metrics, organization and dissemination of datasets, and measuring the effectiveness of current techniques to establish a baseline. We hope rapid progress will soon follow and we look forward to extensive discussion and critical feedback.
\newpage 
\balance
{\small
\bibliographystyle{ieee}
\bibliography{egpaper_final}
}


\end{document}